\documentclass[11pt,a4paper]{article}
\usepackage[hyperref]{acl2019}
\usepackage{times}
\usepackage{latexsym}

\usepackage{url}
\usepackage{helvet}  %Required
\usepackage{courier}  %Required
\usepackage{graphicx}  %Required
\usepackage{multirow}
\usepackage{xspace}
\usepackage{amssymb}
\usepackage{amsmath}
\usepackage{subfigure}
\usepackage{booktabs}
\usepackage{bbm}
\usepackage{array}
\usepackage{bm}
\usepackage[noend]{algpseudocode}
\usepackage{algorithmicx,algorithm}
\usepackage{amssymb}% http://ctan.org/pkg/amssymb
\usepackage{pifont}% http://ctan.org/pkg/pifont
\DeclareMathOperator*{\argmax}{arg\,max}
\graphicspath{{figures/}} % Location of the graphics files

\aclfinalcopy % Uncomment this line for the final submission
\newcommand\bertlarge{BERT$_{\textsc{LARGE}}$\xspace}

\newcommand{\tabincell}[2]{\begin{tabular}{@{}#1@{}}#2\end{tabular}}
\newcommand{\xmark}{\ding{55}}%

\title{Open-Domain Targeted Sentiment Analysis \\ via Span-Based Extraction and Classification}

\author{
Minghao Hu$^\dag$,
Yuxing Peng$^\dag$,
Zhen Huang$^\dag$,
Dongsheng Li$^\dag$,
Yiwei Lv$^\S$ \\
$^\dag$ National University of Defense Technology, Changsha, China \\
$^\S$ University of Macau, Macau, China  \\
{\tt \{huminghao09,pengyuxing,huangzhen,dsli\}@nudt.edu.cn} \\
}

\date{}

\begin{document}
\maketitle

\begin{abstract}

Open-domain targeted sentiment analysis aims to detect opinion targets along with their sentiment polarities from a sentence.
Prior work typically formulates this task as a sequence tagging problem.
However, such formulation suffers from problems such as huge search space and sentiment inconsistency.
To address these problems, we propose a span-based extract-then-classify framework, where multiple opinion targets are directly extracted from the sentence under the supervision of target span boundaries, and corresponding polarities are then classified using their span representations.
We further investigate three approaches under this framework, namely the pipeline, joint, and collapsed models.
Experiments on three benchmark datasets show that our approach consistently outperforms the sequence tagging baseline.
Moreover, we find that the pipeline model achieves the best performance compared with the other two models.

\end{abstract}
\section{Introduction	\label{intro}}
Open-domain targeted sentiment analysis is a fundamental task in opinion mining and sentiment analysis~\cite{pang2008opinion,liu2012sentiment}.
Compared to traditional sentence-level sentiment analysis tasks~\cite{lin2009joint,kim2014convolutional}, the task requires detecting target entities mentioned in the sentence along with their sentiment polarities, thus being more challenging.
Taking Figure \ref{fig:example} as an example, the goal is to first identify ``\emph{Windows 7}'' and ``\emph{Vista}'' as opinion targets and then predict their corresponding sentiment classes.

\begin{figure}[h]
\center
\fbox{\parbox{0.95\columnwidth}{
\begin{small}
\textbf{Sentence:} I love \textbf{[Windows 7]}$_{\text{+}}$ which is a vast improvment over \textbf{[Vista]}$_{\text{-}}$.

\textbf{Targets:} Windows 7, Vista

\textbf{Polarities:} positive, negative
\end{small}
}}
\caption{Open-domain targeted sentiment analysis.}
\label{fig:example}
\end{figure}

Typically, the whole task can be decoupled into two subtasks.
Since opinion targets are not given, we need to first detect the targets from the input text.
This subtask, which is usually denoted as \emph{target extraction}, can be solved by sequence tagging methods~\cite{jakob2010extracting,liu2015fine,wang2016recursive,poria2016aspect,shu2017lifelong,he2017unsupervised,xu2018double}.
Next, \emph{polarity classification} aims to predict the sentiment polarities over the extracted target entities~\cite{jiang2011target,dong2014adaptive,tang2015effective,wang2016attention,chen2017recurrent,xue2018aspect,li2018transformation,fan2018multi}. 
Although lots of efforts have been made to design sophisticated classifiers for this subtask, they all assume that the targets are already given.

\begin{figure*}
  \centering
  \subfigure[Sequence tagging. The B/I/O labels indicate target span boundaries, while +/-/0 refer to sentiment polarities.]{
    \label{scheme:tag} %% label for first subfigure
    \includegraphics[width=.447\textwidth]{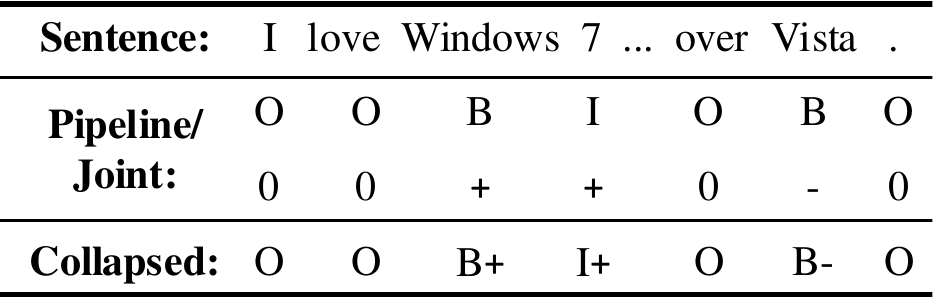}}
  \hspace{0.3in}  
  \subfigure[Span-based labeling. The number denotes the start/end position of the given target in the sentence.]{
    \label{scheme:span} %% label for first subfigure
    \includegraphics[width=.46\textwidth]{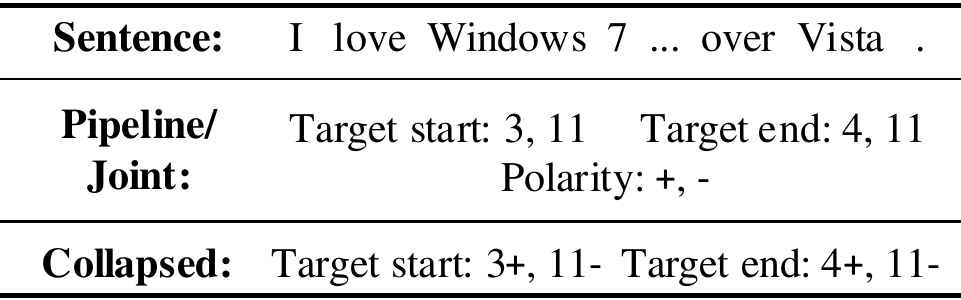}}
  \caption{Comparison of different annotation schemes for the pipeline, joint, and collapsed models.}
  \label{fig:scheme} %% label for entire figure
\end{figure*}

Rather than using separate models for each subtask, some works attempt to solve the task in a more integrated way, by jointly extracting targets and predicting their sentiments~\cite{mitchell2013open,zhang2015neural,li2018unified}.
The key insight is to label each word with a set of target tags (e.g., $\mathrm{B}$, $\mathrm{I}$, $\mathrm{O}$) as well as a set of polarity tags (e.g., +, -, 0), or use a more collapsed set of tags (e.g., $\mathrm{B}$+, $\mathrm{I}$-) to directly indicate the boundary of targeted sentiment, as shown in Figure \ref{scheme:tag}. 
As a result, the entire task is formulated as a sequence tagging problem, and solved using either a pipeline model, a joint model, or a collapsed model under the same network architecture.

However, the above annotation scheme has several disadvantages in target extraction and polarity classification.
\citet{lee2016learning} show that, when using BIO tags for extractive question answering tasks, the model must consider a huge search space due to the compositionality of labels (the power set of all sentence words), thus being less effective.
As for polarity classification, the sequence tagging scheme turns out to be problematic for two reasons. 
First, tagging polarity over each word ignores the semantics of the entire opinion target.
Second, since predicted polarities over target words may be different, the sentiment consistency of multi-word entity can not be guaranteed, as mentioned by \citet{li2018unified}.
For example, there is a chance that the words ``\emph{Windows}'' and ``\emph{7}'' in Figure \ref{scheme:tag} are predicted to have different polarities due to word-level tagging decisions.

To address the problems, we propose a span-based labeling scheme for open-domain targeted sentiment analysis, as shown in Figure \ref{scheme:span}.
The key insight is to annotate each opinion target with its span boundary followed by its sentiment polarity.
Under such annotation, we introduce an extract-then-classify framework that first extracts multiple opinion targets using an heuristic multi-span decoding algorithm, and then classifies their polarities with corresponding summarized span representations.
The advantage of this approach is that the extractive search space can be reduced linearly with the sentence length, which is far less than the tagging method.
Moreover, since the polarity is decided using the targeted span representation, the model is able to take all target words into account before making predictions, thus naturally avoiding sentiment inconsistency.

We take BERT~\cite{devlin2018bert} as the default backbone network, and explore two research questions.
First, we make an elaborate comparison between tagging-based models and span-based models.
Second, following previous works~\cite{mitchell2013open,zhang2015neural}, we compare the pipeline, joint, and collapsed models under the span-based labeling scheme.
Extensive experiments on three benchmark datasets show that our models consistently outperform sequence tagging baselines.
In addition, the pipeline model firmly improves over both the joint and collapsed models.
Source code is released to facilitate future research in this field\footnote{https://github.com/huminghao16/SpanABSA}.
\section{Related Work}
Apart from sentence-level sentiment analysis~\cite{lin2009joint,kim2014convolutional}, targeted sentiment analysis, which requires the detection of sentiments towards mentioned entities in the open domain, is also an important research topic.

As discussed in \S\ref{intro}, this task is usually divided into two subtasks.
The first is target extraction for identifying entities from the input sentence.
Traditionally, Conditional Random Fields (CRF)~\cite{lafferty2001conditional} have been widely explored~\cite{jakob2010extracting,wang2016recursive,shu2017lifelong}.
Recently, many works concentrate on leveraging deep neural networks to tackle this task, e.g., using CNNs~\cite{poria2016aspect,xu2018double}, RNNs~\cite{liu2015fine,he2017unsupervised}, and so on.
The second is polarity classification, assuming that the target entities are given.
Recent works mainly focus on capturing the interaction between the target and the sentence, by utilizing various neural architectures such as LSTMs~\cite{hochreiter1997long,tang2015effective} with attention mechanism~\cite{wang2016attention,li2018transformation,fan2018multi}, CNNs~\cite{xue2018aspect,huang2018parameterized}, and Memory Networks~\cite{tang2016aspect,chen2017recurrent,li2017deep}.

\begin{figure*}
  \centering
  \subfigure[Multi-target extractor.]{
    \label{overview:extra} %% label for first subfigure
    \includegraphics[width=.43\textwidth]{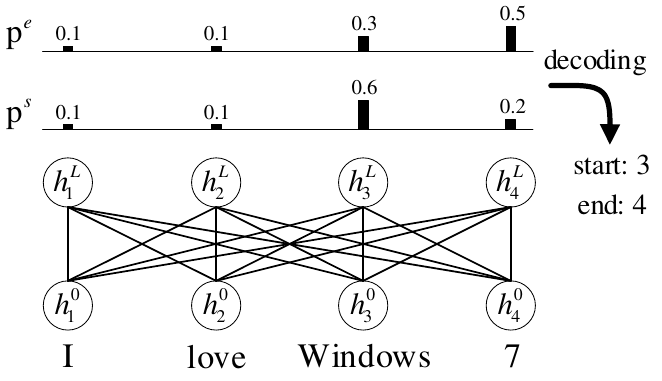}}
  \hspace{0.5in}
  \subfigure[Polarity classifier.]{
    \label{overview:classi} %% label for first subfigure
    \includegraphics[width=.35\textwidth]{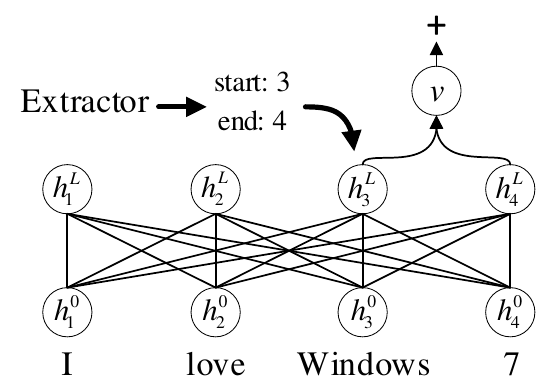}}
  % \subfigure[Collapsed sentiment extractor.]{
  %   \label{overview:coll_extra} %% label for first subfigure
  %   \includegraphics[width=.33\textwidth]{collapsed_extractor}}
  \caption{An overview of the proposed framework. Word embeddings are fed to the BERT encoder~\cite{devlin2018bert} that contains $L$ pre-trained Transformer blocks~\cite{vaswani2017attention}. The last block's hidden states are used to (a) propose one or multiple candidate targets based on the probabilities of the start and end positions, (b) predict the sentiment polarity using the span representation of the given target.}
  % , and (c) produce the collapsed position probability distributions.}
  \label{fig:overview} %% label for entire figure
\end{figure*}

Rather than solving these two subtasks with separate models, a more practical approach is to directly predict the sentiment towards an entity along with discovering the entity itself.
Specifically, \citet{mitchell2013open} formulate the whole task as a sequence tagging problem and propose to use CRF with hand-crafted linguistic features.
\citet{zhang2015neural} further leverage these linguistic features to enhance a neural CRF model.
% Both of these two works investigate the pipeline, joint and collapsed models under the same sequence tagging scheme.
Recently, \citet{li2018unified} have proposed a unified model that contains two stacked LSTMs along with carefully-designed components for maintaining sentiment consistency and improving target word detection.
Our work differs from these approaches in that we formulate this task as a span-level extract-then-classify process instead.

The proposed span-based labeling scheme is inspired by recent advances in machine comprehension and question answering~\cite{seo2016bidirectional,hu2017reinforced}, where the task is to extract a continuous span of text from the document as the answer to the question~\cite{Rajpurkar16}.
To solve this task, \citet{lee2016learning} investigate several predicting strategies, such as BIO prediction, boundary prediction, and the results show that predicting the two endpoints of the answer is more beneficial than the tagging method.
\citet{wang2016machine} explore two answer prediction methods, namely the sequence method and the boundary method, finding that the later performs better.
Our approach is related to this line of work. However, unlike these works that extract one span as the final answer, our approach is designed to dynamically output one or multiple opinion targets.

\section{Extract-then-Classify Framework}
Instead of formulating the open-domain targeted sentiment analysis task as a sequence tagging problem, we propose to use a span-based labeling scheme as follows: 
given an input sentence $\mathbf{x}=(x_1, ..., x_n)$ with length $n$, and a target list $\mathbf{T}=\{\mathbf{t}_1, ..., \mathbf{t}_m\}$, where the number of targets is $m$ and each target $\mathbf{t}_i$ is annotated with its start position, its end position, and its sentiment polarity.
% $\mathbf{y}_i^p \in\mathbb{R}^{3}$ 
The goal is to find all targets from the sentence as well as predict their polarities.

The overall illustration of the proposed framework is shown in Figure \ref{fig:overview}. 
The basis of our framework is the BERT encoder~\cite{devlin2018bert}: we map word embeddings into contextualized token representations using pre-trained Transformer blocks~\cite{vaswani2017attention} (\S\ref{sec:bert}).
A multi-target extractor is first used to propose multiple candidate targets from the sentence (\S\ref{sec:extrac}). 
Then, a polarity classifier is designed to predict the sentiment towards each extracted candidate using its summarized span representation (\S\ref{sec:classi}).
We further investigate three different approaches under this framework, namely the pipeline, joint, and collapsed models in \S\ref{sec:models}.

\subsection{BERT as Backbone Network		\label{sec:bert}}
We use Bidirectional Encoder Representations from Transformers (BERT)~\cite{devlin2018bert}, a pre-trained bidirectional Transformer encoder that achieves state-of-the-art performances across a variety of NLP tasks, as our backbone network.

We first tokenize the sentence $\mathbf{x}$ using a 30,522 wordpiece vocabulary, and then generate the input sequence $\mathbf{\tilde{x}}$ by concatenating a \texttt{[CLS]} token, the tokenized sentence, and a \texttt{[SEP]} token.
Then for each token $\tilde{x}_i$ in $\mathbf{\tilde{x}}$, we convert it into vector space by summing the token, segment, and position embeddings, thus yielding the input embeddings $\mathbf{h}^0 \in\mathbb{R}^{(n+2) \times h}$, where $h$ is the hidden size.

Next, we use a series of $L$ stacked Transformer blocks to project the input embeddings into a sequence of contextual vectors $\mathbf{h}^i \in\mathbb{R}^{(n+2) \times h}$ as:
\begin{equation}
\mathbf{h}^i = \mathrm{TransformerBlock}(\mathbf{h}^{i-1}), \forall i \in [1,L]	\nonumber
\end{equation}
Here, we omit an exhaustive description of the block architecture and refer readers to \citet{vaswani2017attention} for more details.

\subsection{Multi-Target Extractor		\label{sec:extrac}}
Multi-target extractor aims to propose multiple candidate opinion targets (Figure \ref{overview:extra}).
Rather than finding targets via sequence tagging methods, we detect candidate targets by predicting the start and end positions of the target in the sentence, as suggested in extractive question answering~\cite{wang2016machine,seo2016bidirectional,hu2017reinforced}.
We obtain the unnormalized score as well as the probability distribution of the start position as:
\begin{equation}
\mathbf{g}^s = \mathbf{w}_s \mathbf{h}^L \ , \  \mathbf{p}^s = \mathrm{softmax}(\mathbf{g}^s)	\nonumber \\
\end{equation}
where $\mathbf{w}_s \in\mathbb{R}^{h}$ is a trainable weight vector. Similarly, we can get the probability of the end position along with its confidence score by:
\begin{equation}
\mathbf{g}^e = \mathbf{w}_e \mathbf{h}^L \ , \  \mathbf{p}^e = \mathrm{softmax}(\mathbf{g}^e)	\nonumber \\
\end{equation}

During training, since each sentence may contain multiple targets, we label the span boundaries for all target entities in the list $\mathbf{T}$.
As a result, we can obtain a vector $\mathbf{y}^s \in\mathbb{R}^{(n+2)}$, where each element $\mathbf{y}^s_i$ indicates whether the $i$-th token starts a target, and also get another vector $\mathbf{y}^e \in\mathbb{R}^{(n+2)}$ for labeling the end positions.
Then, we define the training objective as the sum of the negative log probabilities of the true start and end positions on two predicted probabilities as:
\begin{equation} 
	\mathcal{L} = - \sum\nolimits_{i=1}^{n+2} \mathbf{y}^s_i \log (\mathbf{p}^s_i) - \sum\nolimits_{j=1}^{n+2} \mathbf{y}^e_j \log (\mathbf{p}^e_j)	\nonumber
\end{equation}

At inference time, previous works choose the span $(k, l)$ ($k \le l$) with the maximum value of $\mathbf{g}^s_k + \mathbf{g}^e_l$ as the final prediction.
However, such decoding method is not suitable for the multi-target extraction task.
Moreover, simply taking top-$K$ spans according to the addition of two scores is also not optimal, as multiple candidates may refer to the same text. 
Figure \ref{fig:redun} gives a qualitative example to illustrate this phenomenon. 

\begin{figure}[h]
\center
\fbox{\parbox{0.95\columnwidth}{
\begin{small}
\textbf{Sentence:} Great food but the service was dreadful!

\textbf{Targets:} food, service

\textbf{Predictions:} food but the service, food, Great food, service, service was dreadful, ...
\end{small}
}}
\caption{An example shows that there are many redundant spans in top-$K$ predictions.}
\label{fig:redun}
\end{figure}

To adapt to multi-target scenarios, we propose an heuristic multi-span decoding algorithm as shown in Algorithm \ref{algo:hmsd}.
For each example, top-$M$ indices are first chosen from the two predicted scores $\mathbf{g}^s$ and $\mathbf{g}^e$ (line 2), and the candidate span $(\mathbf{s}_i, \mathbf{e}_j)$ (denoted as $\mathbf{r}_l$) along with its heuristic-regularized score $\mathbf{u}_l$ are then added to the lists $\mathbf{R}$ and $\mathbf{U}$ respectively, under the constraints that the end position is no less than the start position as well as the addition of two scores exceeds a threshold $\gamma$ (line 3-8).
Note that we heuristically calculate $\mathbf{u}_l$ as the sum of two scores minus the span length (line 6), which turns out to be critical to the performance as targets are usually short entities.
Next, we prune redundant spans in $\mathbf{R}$ using the non-maximum suppression algorithm~\cite{rosenfeld1971edge}.
Specifically, we remove the span $\mathbf{r}_l$ that possesses the maximum score $\mathbf{u}_l$ from the set $\mathbf{R}$ and add it to the set $\mathbf{O}$ (line 10-11).
We also delete any span $\mathbf{r}_k$ that is overlapped with $\mathbf{r}_l$, which is measured with the word-level F1 function (line 12-14).
This process is repeated for remaining spans in $\mathbf{R}$, until $\mathbf{R}$ is empty or top-$K$ target spans have been proposed (line 9).

\begin{algorithm}[h]
\small
\caption{Heuristic multi-span decoding} %算法的名字
\label{algo:hmsd}
{\bf Input:} %算法的输入， \hspace*{0.02in}用来控制位置，同时利用 \\ 进行换行
$\mathbf{g}^s$, $\mathbf{g}^e$, $\gamma$, $K$ \\
\hspace*{0.15in} $\mathbf{g}^s$ denotes the score of start positions \\
\hspace*{0.15in} $\mathbf{g}^e$ denotes the score of end positions \\
\hspace*{0.15in} $\gamma$ is a minimum score threshold \\
\hspace*{0.15in} $K$ is the maximum number of proposed targets
\begin{algorithmic}[1]
\State Initialize $\mathbf{R}, \mathbf{U}, \mathbf{O} = \{\}, \{\}, \{\}$ % \State 后写一般语句
\State Get top-$M$ indices $\mathbf{S}$, $\mathbf{E}$ from $\mathbf{g}^s$, $\mathbf{g}^e$ 

\For{$\mathbf{s}_i$ in $\mathbf{S}$} % For 语句，需要和EndFor对应
	\For{$\mathbf{e}_j$ in $\mathbf{E}$} 
	　　\If{$\mathbf{s}_i \le \mathbf{e}_j$ and $\mathbf{g}^s_{\mathbf{s}_i} + \mathbf{g}^e_{\mathbf{e}_j} \ge \gamma$}
		\State $\mathbf{u}_l = \mathbf{g}^s_{\mathbf{s}_i} + \mathbf{g}^e_{\mathbf{e}_j} - (\mathbf{e}_j - \mathbf{s}_i + 1)$
		\State $\mathbf{r}_l = (\mathbf{s}_i, \mathbf{e}_j)$
		\State $\mathbf{R} = \mathbf{R} \cup \{\mathbf{r}_l\}$, $\mathbf{U} = \mathbf{U} \cup \{\mathbf{u}_l\}$
		\EndIf
	\EndFor	　　
\EndFor	

\While{$\mathbf{R} \ne \{\}$ and $\mathrm{size}(\mathbf{O}) < K$} % While语句，需要和EndWhile对应
　　\State $l = \argmax \mathbf{U}$
	\State $\mathbf{O} = \mathbf{O} \cup \{\mathbf{r}_l\}$; $\mathbf{R} = \mathbf{R} - \{\mathbf{r}_l\}$; $\mathbf{U} = \mathbf{U} - \{\mathbf{u}_l\}$

	\For{$\mathbf{r}_k$ in $\mathbf{R}$} % For 语句，需要和EndFor对应
	　　\If{$\mathrm{f1}(\mathbf{r}_l, \mathbf{r}_k) \ne 0$} % If 语句，需要和EndIf对应
	　　　　\State $\mathbf{R} = \mathbf{R} - \{\mathbf{r}_k\}$; $\mathbf{U} = \mathbf{U} - \{\mathbf{u}_k\}$
	　　\EndIf
	\EndFor

\EndWhile
\State \Return $\mathbf{O}$
\end{algorithmic}
\end{algorithm}

\subsection{Polarity Classifier		\label{sec:classi}}
Typically, polarity classification is solved using either sequence tagging methods or sophisticated neural networks that separately encode the target and the sentence.
Instead, we propose to summarize the target representation from contextual sentence vectors according to its span boundary, and use feed-forward neural networks to predict the sentiment polarity, as shown in Figure \ref{overview:classi}.

Specifically, given a target span $\mathbf{r}$, we calculate a summarized vector $\mathbf{v}$ using the attention mechanism~\cite{bahdanau2014neural} over tokens in its corrsponding bound $(\mathbf{s}_i, \mathbf{e}_j)$, similar to \citet{lee2017end} and \citet{he2018jointly}:
\begin{gather}
\alpha = \mathrm{softmax}(\mathbf{w}_{\alpha} \mathbf{h}^L_{\mathbf{s}_i:\mathbf{e}_j}) \nonumber \\
\mathbf{v} = \sum\nolimits_{t=\mathbf{s}_i}^{\mathbf{e}_j} {\alpha_{t-\mathbf{s}_i+1}} \mathbf{h}_t^L  \nonumber
\end{gather}
where $\mathbf{w}_{\alpha} \in\mathbb{R}^{h}$ is a trainable weight vector.

The polarity score is obtained by applying two linear transformations with a Tanh activation in between, and is normalized with the softmax function to output the polarity probability as:
\begin{equation} 
	\mathbf{g}^p = \mathbf{W}_{p} \mathrm{tanh}(\mathbf{W}_{v}  \mathbf{v}) \ , \  \mathbf{p}^p = \mathrm{softmax}(\mathbf{g}^p)  \nonumber
\end{equation}
where $\mathbf{W}_{v} \in\mathbb{R}^{h \times h}$ and $\mathbf{W}_{p} \in\mathbb{R}^{k \times h}$ are two trainable parameter matrices.

We minimize the negative log probabilities of the true polarity on the predicted probability as:
\begin{equation} 
	\mathcal{J} = - \sum\nolimits_{i=1}^{k} \mathbf{y}^p_i \log ( \mathbf{p}^p_i )	\nonumber
\end{equation}
where $\mathbf{y}^p$ is an one-hot label indicating the true polarity, and $k$ is the number of sentiment classes.

During inference, the polarity probability is calculated for each candidate target span in the set $\mathbf{O}$, and the sentiment class that possesses the maximum value in $\mathbf{p}^p$ is chosen.

\subsection{Model Variants		\label{sec:models}}
Following \citet{mitchell2013open,zhang2015neural}, we investigate three kinds of models under the extract-then-classify framework:

\paragraph{Pipeline model}
We first build a multi-target extractor where a BERT encoder is exclusively used. 
Then, a second backbone network is used to provide contextual sentence vectors for the polarity classifier.
Two models are separately trained and combined as a pipeline during inference.
% The classifier takes the proposed candidate targets from the extractor and outputs predicted polarities.

\paragraph{Joint model} 
In this model, each sentence is fed into a shared BERT backbone network that finally branches into two sibling output layers: one for proposing multiple candidate targets and another for predicting the sentiment polarity over each extracted target.
A joint training loss $\mathcal{L} + \mathcal{J}$ is used to optimize the whole model.
The inference procedure is the same as the pipeline model.

\paragraph{Collapsed model}
We combine target span boundaries and sentiment polarities into one label space. 
For example, the sentence in Figure \ref{scheme:span} has a positive span $(3\text{+}, 4\text{+})$ and a negative span $(11\text{-}, 11\text{-})$.
We then modify the multi-target extractor by producing three sets of probabilities of the start and end positions, where each set corresponds to one sentiment class ( e.g., $\mathbf{p}^{s+}$ and $\mathbf{p}^{e+}$ for positive targets).
Then, we define three objectives to optimize towards each polarity.
During inference, the heuristic multi-span decoding algorithm is performed on each set of scores (e.g., $\mathbf{g}^{s+}$ and $\mathbf{g}^{e+}$), and the output sets $\mathbf{O}^+$, $\mathbf{O}^-$, and $\mathbf{O}^0$ are aggregated as the final prediction.
\section{Experiments}

\subsection{Setup}

\begin{table}
\begin{center}
\resizebox{0.95\columnwidth}{!}{
\begin{tabular}{l|ccccc}
\toprule
Dataset  & \#Sent  & \#Targets & \#+ & \#- & \#0 \\ 
\midrule
\texttt{LAPTOP}     & 1,869 & 2,936 & 1,326 & 990 & 620 \\
\texttt{REST}    & 3,900 & 6,603 & 4,134 & 1,538 & 931 \\
\texttt{TWITTER}         & 2,350 & 3,243 & 703 & 274 & 2,266 \\
\bottomrule
\end{tabular}}
\caption{\label{table:data} Dataset statistics. `\#Sent' and `\#Targets' denote the number of sentences and targets, respectively. `+', `-', and `0' refer to the positive, negative, and neutral sentiment classes.}
\end{center}
\end{table}

\begin{table*}[]
    \centering
    \begin{tabular}{l|ccc|ccc|ccc}
    \toprule
    \multirow{2}*{ Model } & \multicolumn{3}{c}{\texttt{LAPTOP}} & \multicolumn{3}{c}{\texttt{REST}} & \multicolumn{3}{c}{\texttt{TWITTER}}\\ 
    % \cline{3-11} 
     & Prec. & Rec. & F1 & Prec. & Rec. & F1 & Prec. & Rec. & F1 \\
    \midrule
    \midrule
    UNIFIED        & 61.27 & 54.89 & 57.90 & 68.64 & 71.01 & 69.80 & 53.08 & 43.56 & 48.01 \\
    \midrule
    TAG-pipeline   & 65.84 & 67.19 & 66.51 & 71.66 & 76.45 & 73.98 & 54.24 & 54.37 & 54.26 \\
    TAG-joint      & 65.43 & 66.56 & 65.99 & 71.47 & 75.62 & 73.49 & 54.18 & 54.29 & 54.20 \\ 
    TAG-collapsed  & 63.71 & 66.83 & 65.23 & 71.05 & 75.84 & 73.35 & 54.05 & 54.25 & 54.12 \\
    \midrule
    SPAN-pipeline  & 69.46 & 66.72 & \textbf{68.06} & 76.14 & 73.74 & \textbf{74.92} & 60.72 & 55.02 & \textbf{57.69} \\
    SPAN-joint     & 67.41 & 61.99 & 64.59 & 72.32 & 72.61 & 72.47 & 57.03 & 52.69 & 54.55 \\ 
    SPAN-collapsed & 50.08 & 47.32 & 48.66 & 63.63 & 53.04 & 57.85 & 51.89 & 45.05 & 48.11 \\
    \bottomrule
    \end{tabular}
    \caption{Main results on three benchmark datasets. A \bertlarge backbone network is used for both the ``TAG'' and ``SPAN'' models. State-of-the-art results are marked in \textbf{bold}.}
    \label{tab:main_results}
\end{table*}

\paragraph{Datasets}
We conduct experiments on three benchmark sentiment analysis datasets, as shown in Table \ref{table:data}.
\texttt{LAPTOP} contains product reviews from the laptop domain in SemEval 2014 ABSA challenges~\cite{pontiki2014semeval}.
\texttt{REST} is the union set of the restaurant domain from SemEval 2014, 2015 and 2016~\cite{pontiki2015semeval,pontiki2016semeval}.
\texttt{TWITTER} is built by \citet{mitchell2013open}, consisting of twitter posts.
Following \citet{zhang2015neural,li2018unified}, we report the ten-fold cross validation results for \texttt{TWITTER}, as there is no train-test split.
For each dataset, the gold target span boundaries are available, and the targets are labeled with three sentiment polarities, namely \emph{positive} (+), \emph{negative} (-), and \emph{neutral} (0).

\paragraph{Metrices}
We adopt the precision (P), recall (R), and F1 score as evaluation metrics. 
A predicted target is correct only if it exactly matches the gold target entity and the corresponding polarity.
To separately analyze the performance of two subtasks, precision, recall, and F1 are also used for the target extraction subtask, while the accuracy (ACC) metric is applied to polarity classification.

\paragraph{Model settings}
We use the publicly available \bertlarge\footnote{https://github.com/google-research/bert} model as our backbone network, and refer readers to \citet{devlin2018bert} for details on model sizes.
We use Adam optimizer with a learning rate of 2e-5 and warmup over the first 10\% steps to train for 3 epochs.
The batch size is 32 and a dropout probability of 0.1 is used.
The number of candidate spans $M$ is set as 20 while the maximum number of proposed targets $K$ is 10 (Algorithm \ref{algo:hmsd}).
The threshold $\gamma$ is manually tuned on each dataset.
All experiments are conducted on a single NVIDIA P100 GPU card.

\subsection{Baseline Methods}
We compare the proposed span-based approach with the following methods:

\noindent{\textbf{TAG}-\{\textbf{pipeline}, \textbf{joint}, \textbf{collapsed}\}} are the sequence tagging baselines that involve a BERT encoder and a CRF decoder. ``pipeline'' and ``joint'' denote the pipeline and joint approaches that utilize the BIO and +/-/0 tagging schemes, while ``collapsed'' is the model following the collapsed tagging scheme (Figure \ref{scheme:tag}).

\noindent{\textbf{UNIFIED}}~\cite{li2018unified} is the current state-of-the-art model on targeted sentiment analysis\footnote{https://github.com/lixin4ever/E2E-TBSA}. It contains two stacked recurrent neural networks enhanced with multi-task learning and adopts the collapsed tagging scheme.

We also compare our multi-target extractor with the following method:

\noindent{\textbf{DE-CNN}}~\cite{xu2018double} is the current state-of-the-art model on target extraction, which combines a double embeddings mechanism with convolutional neural networks (CNNs)\footnote{https://www.cs.uic.edu/˜hxu/}.

Finally, the polarity classifier is compared with the following methods:

\noindent{\textbf{MGAN}}~\cite{fan2018multi} uses a multi-grained attention mechanism to capture interactions between targets and sentences for polarity classification.

\noindent{\textbf{TNet}}~\cite{li2018transformation} is the current state-of-the-art model on polarity classification, which consists of a multi-layer context-preserving network architecture and uses CNNs as feature extractor\footnote{https:// github.com/lixin4ever/TNet}.

\subsection{Main Results}
We compare models under either the sequence tagging scheme or the span-based labeling scheme, and show the results in Table \ref{tab:main_results}.
We denote our approach as ``SPAN'', and use \bertlarge as backbone networks for both the ``TAG'' and ``SPAN'' models to make the comparison fair.

Two main observations can be obtained from the Table.
First, despite that the ``TAG'' baselines already outperform previous best approach (``UNIFIED''), they are all beaten by the ``SPAN'' methods.
The best span-based method achieves 1.55\%, 0.94\% and 3.43\% absolute gains on three datasets compared to the best tagging method, indicating the efficacy of our extract-then-classify framework.
Second, among the span-based methods, the SPAN-pipeline achieves the best performance, which is similar to the results of \citet{mitchell2013open,zhang2015neural}. 
This suggests that there is only a weak connection between target extraction and polarity classification.
The conclusion is also supported by the result of SPAN-collapsed method, which severely drops across all datasets, implying that merging polarity labels into target spans does not address the task effectively.

\begin{table}
\begin{center}
% \small
\begin{tabular}{l|c|c|c}
\toprule
Model  & \texttt{LAPTOP}  & \texttt{REST} & \texttt{TWITTER} \\ 
\midrule
DE-CNN  & 81.59 & - & -  \\
TAG     & \textbf{85.20} & \textbf{84.48} & 73.47  \\
SPAN    & 83.35 & 82.38 & \textbf{75.28}  \\
% w/o heuristics    & 81.95 & 74.06 & 72.31  \\
% w/o NMS	& 68.33 & 68.09 & 62.89 \\
\bottomrule
\end{tabular}
\caption{\label{tab:target_extra} F1 comparison of different approaches for target extraction.}
\end{center}
\end{table}

\begin{figure}
\center
\includegraphics[width=.95\columnwidth]{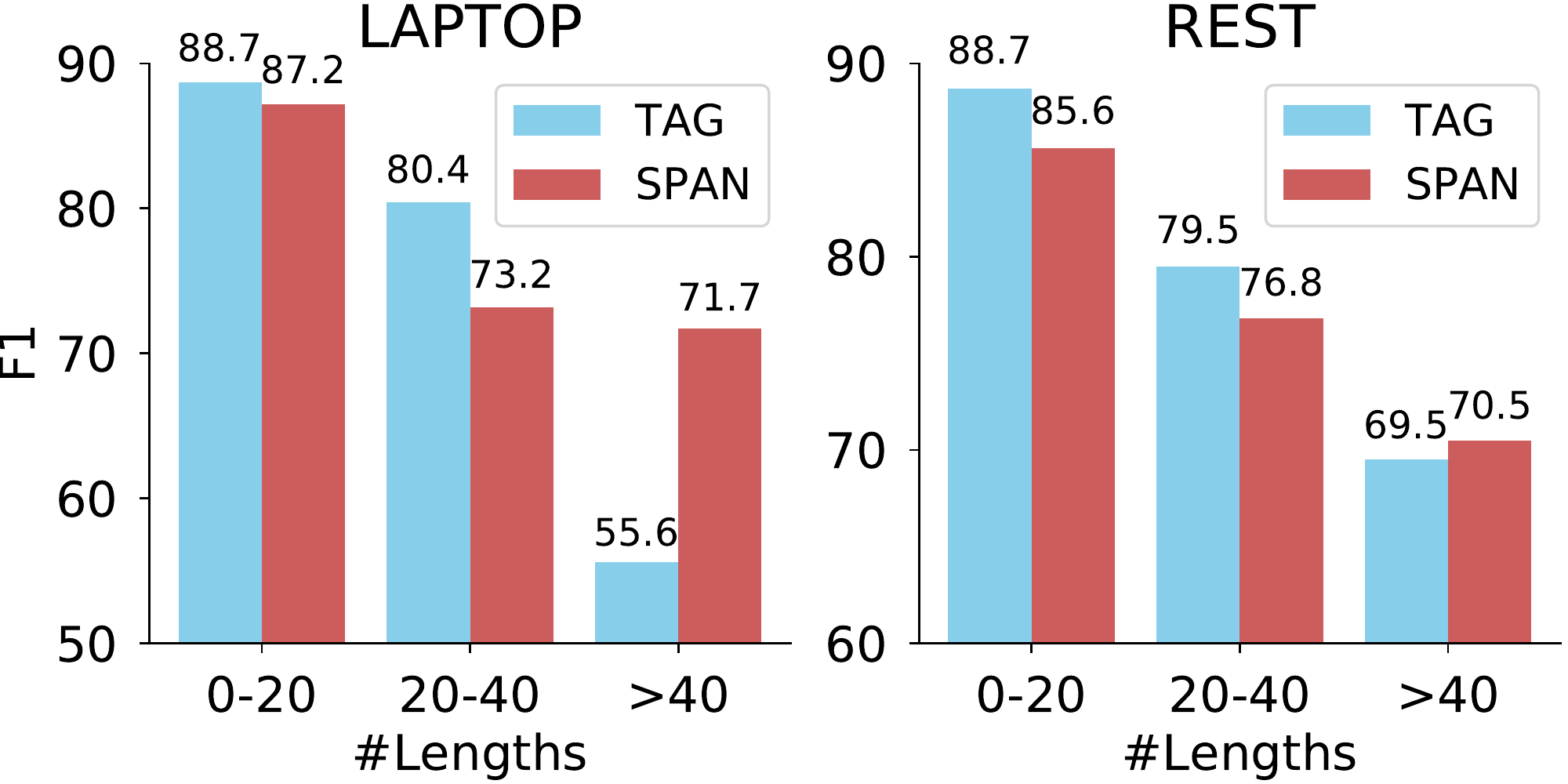}
\caption{F1 on \texttt{LAPTOP} and \texttt{REST} w.r.t different sentence lengths for target extraction.}
\label{fig:seq_len}
\end{figure}

\subsection{Analysis on Target Extraction}
To analyze the performance on target extraction, we run both the tagging baseline and the multi-target extractor on three datasets, as shown in Table \ref{tab:target_extra}.
We find that the BIO tagger outperforms our extractor on \texttt{LAPTOP} and \texttt{REST}.
A likely reason for this observation is that the lengths of input sentences on these datasets are usually small (e.g., 98\% of sentences are less than 40 words in \texttt{REST}), which limits the tagger's search space (the power set of all sentence words).
As a result, the computational complexity has been largely reduced, which is beneficial for the tagging method.

In order to confirm the above hypothesis, we plot the F1 score with respect to different sentence lengths in Figure \ref{fig:seq_len}. 
We observe that the performance of BIO tagger dramatically decreases as the sentence length increases, while our extractor is more robust for long sentences.
Our extractor manages to surpass the tagger by 16.1 F1 and 1.0 F1 when the length exceeds 40 on \texttt{LAPTOP} and \texttt{REST}, respectively.
The above result demonstrates that our extractor is more suitable for long sentences due to the fact that its search space only increases linearly with the sentence length.
% our extractor gives higher F1 scores than the BIO tagger on \texttt{LAPTOP} and \texttt{TWITTER} datasets, while provides comparable results on the \texttt{REST} dataset.
% The gains on \texttt{LAPTOP} and \texttt{TWITTER} are mostly attributed to improved precisions, while the recall of our extractor are relatively lower.
% A likely reson for this observation is that a relatively larger threshold $\gamma$ has been set, which allows less candidate targets to be proposed by our decoding algorithm.
% As a result, the model only makes cautious but confident predictions.
% In contrast, the tagging method does not rely on the threshold and is observed to have a higher recall than the precision.

\begin{figure}
\center
\includegraphics[width=.95\columnwidth]{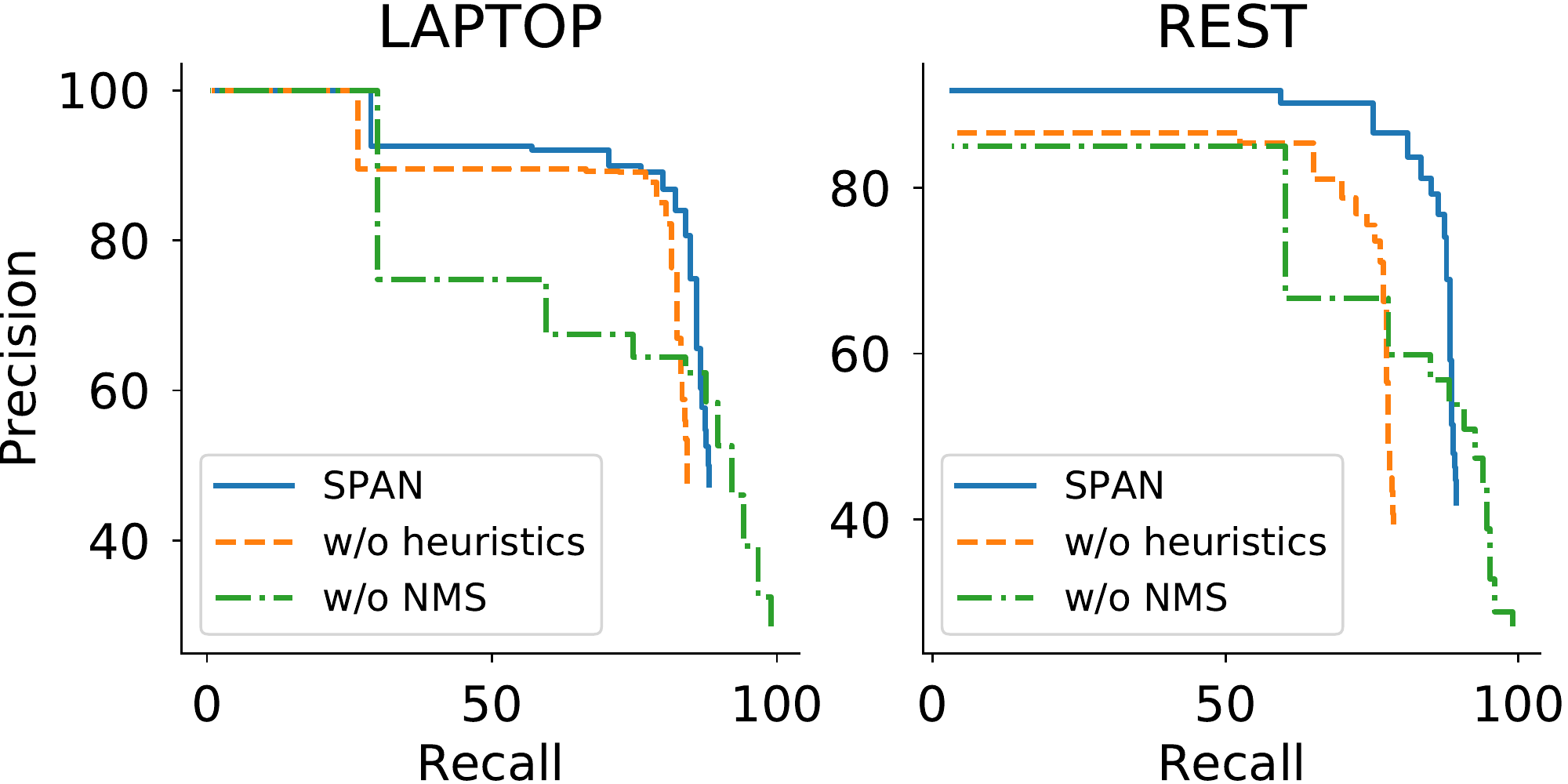}
\caption{Precision-recall curves on \texttt{LAPTOP} and \texttt{REST} for target extraction.  ``NMS'' and ``heuristics'' denote the non-maximum suppression and the length heuristics in Algorithm \ref{algo:hmsd}.}
\label{fig:pr_curve}
\end{figure}

Since a trade-off between precision and recall can be adjusted according to the threshold $\gamma$ in our extractor, we further plot the precision-recall curves under different ablations to show the effects of heuristic multi-span decoding algorithm.
As can be seen from Figure \ref{fig:pr_curve}, ablating the length heuristics results in consistent performance drops across two datasets.
By sampling incorrect predictions we find that there are many targets closely aligned with each other, such as ``\emph{perfect \textbf{[size]}$_{\text{+}}$ and \textbf{[speed]}$_{\text{+}}$}'', ``\emph{\textbf{[portions]}$_{\text{+}}$ all at a reasonable \textbf{[price]}$_{\text{+}}$}'', and so on.
The model without length heuristics is very likely to output the whole phrase as a single target, thus being totally wrong.
Moreover, removing the non-maximum suppression (NMS) leads to significant performance degradations, suggesting that it is crucial to prune redundant spans that refer to the same text.

\begin{table*}[]
    \centering
    \resizebox{0.99\textwidth}{!}{
    \begin{tabular}{l|c|c}
    \toprule
    Sentence & TAG & SPAN \\ 
    \midrule
    \midrule
    \tabincell{l}{1. I thought the transition would be difficult at best and would take some time \\ to fully familiarize myself with the new \textbf{[Mac ecosystem]$_{\text{0}}$}.} & [ecosystem]$_{\text{+}}$ (\xmark) & [Mac ecosystem]$_{\text{0}}$ \\
    \midrule
    \tabincell{l}{2. I would normally not finish the \textbf{[brocolli]$_{\text{+}}$} when I order these kinds of food \\ but for the first time, every piece was as eventful as the first one... the \textbf{[scallops]$_{\text{+}}$} \\ and \textbf{[prawns]$_{\text{+}}$} was so fresh and nicely cooked.} & \tabincell{c}{[brocolli]$_{\text{-}}$ (\xmark), \\ {[scallops and prawns]}$_{\text{+}}$ (\xmark), \\ {[food]}$_{\text{0}}$ (\xmark) }  & \tabincell{c}{ [brocolli]$_{\text{+}}$, \\ {[scallops]}$_{\text{+}}$, \\ {[prawns]}$_{\text{+}}$ } \\
    \midrule
	\tabincell{l}{3. I like the \textbf{[brightness]$_{\text{+}}$} and \textbf{[adjustments]$_{\text{+}}$}.} & [brightness]$_{\text{+}}$, [adjustments]$_{\text{+}}$ & [brightness]$_{\text{+}}$, None (\xmark) \\
    \midrule
	\tabincell{l}{4. The \textbf{[waiter]$_{\text{-}}$} was a bit unfriendly and the \textbf{[feel]$_{\text{-}}$} of the restaurant was crowded.} & [waiter]$_{\text{-}}$, [feel]$_{\text{-}}$ & [waiter]$_{\text{-}}$, None (\xmark) \\
	\midrule
	\tabincell{l}{5. However, it did not have any scratches, zero \textbf{[battery cycle count]$_{\text{+}}$} (pretty \\ surprised), and all the \textbf{[hardware]$_{\text{+}}$} seemed to be working perfectly.} &  \tabincell{c}{[battery cycle count]$_{\text{0}}$ (\xmark), \\ {[hardware]}$_{\text{+}}$ } & \tabincell{c}{[battery cycle count]$_{\text{+}}$, \\ {[hardware]}$_{\text{+}}$ } \\
    \midrule    
    \tabincell{l}{6. I agree that dining at \textbf{[Casa La Femme]$_{\text{-}}$} is like no other dining experience!} & [Casa La Femme]$_{\text{+}}$ (\xmark) & [Casa La Femme]$_{\text{-}}$ \\
    \bottomrule
    \end{tabular}}
    \caption{Case study. The extracted targets are wrapped in brackets with the predicted polarities given as subscripts. Incorrect predictions are marked with \xmark.}
    \label{tab:case_study}
\end{table*}

\subsection{Analysis on Polarity Classification}
To assess the polarity classification subtask, we compare the performance of our span-level polarity classifier with the CRF-based tagger in Table \ref{table:polarity_classi}. 
The results show that our approach significantly outperforms the tagging baseline by achieving 9.97\%, 8.15\% and 15.4\% absolute gains on three datasets, and firmly surpasses previous state-of-the-art models on \texttt{LAPTOP}.
The large improvement over the tagging baseline suggests that detecting sentiment with the entire span representation is much more beneficial than predicting polarities over each word, as the semantics of the given target has been fully considered. 

\begin{figure}
\center
\includegraphics[width=.95\columnwidth]{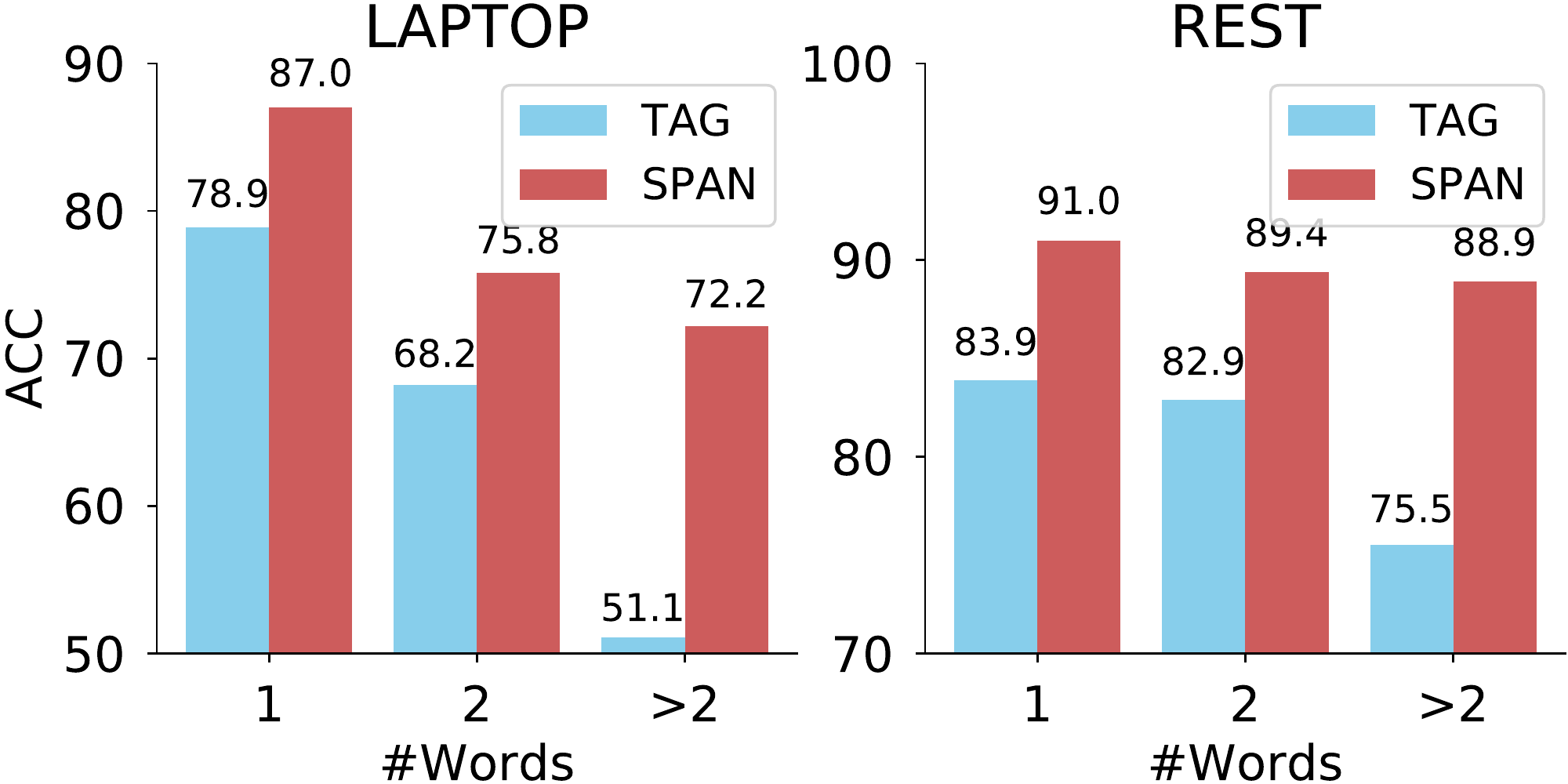}
\caption{Accuracy on \texttt{LAPTOP} and \texttt{REST} w.r.t different number of target words for polarity classification.}
\label{fig:num_word}
\end{figure}

To gain more insights on performance improvements, we plot the accuracy of both methods with respect to different target lengths in Figure \ref{fig:num_word}.
We find that the accuracy of span-level classifier only drops a little as the number of words increases on the \texttt{LAPTOP} and \texttt{REST} datasets.
The performance of tagging baseline, however, significantly decreases as the target becomes longer.
It demonstrates that the tagging method indeed suffers from the sentiment inconsistency problem when it comes to multi-word target entities.
Our span-based method, on the contrary, can naturally alleviate such problem because the polarity is classified by taking all target words into account.

\begin{table}
\begin{center}
% \small
\begin{tabular}{l|c|c|c}
\toprule
Model  & \texttt{LAPTOP}  & \texttt{REST} & \texttt{TWITTER} \\ 
\midrule
MGAN    & 75.39 & - & -  \\
TNet    & 76.54 & - & -  \\
TAG     & 71.42 & 81.80 & 59.76  \\
SPAN    & \textbf{81.39} & \textbf{89.95} & \textbf{75.16}  \\
\bottomrule
\end{tabular}
\caption{\label{table:polarity_classi} Accuracy comparison of different approaches for polarity classification.}
\end{center}
\end{table}

\subsection{Case Study}
Table \ref{tab:case_study} shows some qualitative cases sampled from the pipeline methods.
As observed in the first two examples, the ``TAG'' model incorrectly predicts the target span by either missing the word ``\emph{Mac}'' or proposing a phrase across two targets (``\emph{scallps and prawns}'').
A likely reason of its failure is that the input sentences are relatively longer, and the tagging method is less effective when dealing with them.
But when it comes to shorter inputs (e.g., the third and the fourth examples), the tagging baseline usually performs better than our approach.
We find that our approach may sometimes fail to propose target entities (e.g., ``\emph{adjustments}'' in (3) and ``\emph{feel}'' in (4)), which is due to the fact that a relatively large $\gamma$ has been set.
As a result, the model only makes cautious but confident predictions.
% , and therefore has a high precision but a low recall.
In contrast, the tagging method does not rely on a threshold and is observed to have a higher recall.
For example, it additionally predicts the entity ``\emph{food}'' as a target in the second example.
Moreover, we find that the tagging method sometimes fails to predict the correct sentiment class, especially when the target consists of multiple words (e.g., ``\emph{battery cycle count}'' in (5) and ``\emph{Casa La Femme}'' in (6)), indicating the tagger can not effectively maintain sentiment consistency across words.
% that it is difficult to maintain sentiment consistency for multi-word target.
Our polarity classifier, however, can avoid such problem by using the target span representation to predict the sentiment.

\section{Conclusion}
We re-examine the drawbacks of sequence tagging methods in open-domain targeted sentiment analysis, and propose an extract-then-classify framework with the span-based labeling scheme instead.
The framework contains a pre-trained Transformer encoder as the backbone network.
On top of it, we design a multi-target extractor for proposing multiple candidate targets with an heuristic multi-span decoding algorithm, and introduce a polarity classifier that predicts the sentiment towards each candidate using its summarized span representation.
Our approach firmly outperforms the sequence tagging baseline as well as previous state-of-the-art methods on three benchmark datasets.
Model analysis reveals that the main performance improvement comes from the span-level polarity classifier, and the multi-target extractor is more suitable for long sentences.
Moreover, we find that the pipeline model consistently surpasses both the joint model and the collapsed model.

\section*{Acknowledgments}
We thank the anonymous reviewers for their insightful feedback.
We also thank Li Dong for his helpful comments and suggestions.
This work was supported by the National Key Research and Development Program of China
(2016YFB1000101).

\bibliography{sections/reference}

\begin{thebibliography}{41}
\expandafter\ifx\csname natexlab\endcsname\relax\def\natexlab#1{#1}\fi

\bibitem[{Bahdanau et~al.(2014)Bahdanau, Cho, and Bengio}]{bahdanau2014neural}
Dzmitry Bahdanau, Kyunghyun Cho, and Yoshua Bengio. 2014.
\newblock Neural machine translation by jointly learning to align and
  translate.
\newblock \emph{arXiv preprint arXiv:1409.0473}.

\bibitem[{Chen et~al.(2017)Chen, Sun, Bing, and Yang}]{chen2017recurrent}
Peng Chen, Zhongqian Sun, Lidong Bing, and Wei Yang. 2017.
\newblock Recurrent attention network on memory for aspect sentiment analysis.
\newblock In \emph{Proceedings of EMNLP}.

\bibitem[{Devlin et~al.(2018)Devlin, Chang, Lee, and
  Toutanova}]{devlin2018bert}
Jacob Devlin, Ming-Wei Chang, Kenton Lee, and Kristina Toutanova. 2018.
\newblock Bert: Pre-training of deep bidirectional transformers for language
  understanding.
\newblock \emph{arXiv preprint arXiv:1810.04805}.

\bibitem[{Dong et~al.(2014)Dong, Wei, Tan, Tang, Zhou, and
  Xu}]{dong2014adaptive}
Li~Dong, Furu Wei, Chuanqi Tan, Duyu Tang, Ming Zhou, and Ke~Xu. 2014.
\newblock Adaptive recursive neural network for target-dependent twitter
  sentiment classification.
\newblock In \emph{Proceedings of ACL}.

\bibitem[{Fan et~al.(2018)Fan, Feng, and Zhao}]{fan2018multi}
Feifan Fan, Yansong Feng, and Dongyan Zhao. 2018.
\newblock Multi-grained attention network for aspect-level sentiment
  classification.
\newblock In \emph{Proceedings of EMNLP}.

\bibitem[{He et~al.(2018)He, Lee, Levy, and Zettlemoyer}]{he2018jointly}
Luheng He, Kenton Lee, Omer Levy, and Luke Zettlemoyer. 2018.
\newblock Jointly predicting predicates and arguments in neural semantic role
  labeling.
\newblock \emph{arXiv preprint arXiv:1805.04787}.

\bibitem[{He et~al.(2017)He, Lee, Ng, and Dahlmeier}]{he2017unsupervised}
Ruidan He, Wee~Sun Lee, Hwee~Tou Ng, and Daniel Dahlmeier. 2017.
\newblock An unsupervised neural attention model for aspect extraction.
\newblock In \emph{Proceedings of ACL}.

\bibitem[{Hochreiter and Schmidhuber(1997)}]{hochreiter1997long}
Sepp Hochreiter and J{\"u}rgen Schmidhuber. 1997.
\newblock Long short-term memory.
\newblock \emph{Neural computation}.

\bibitem[{Hu et~al.(2018)Hu, Peng, Huang, Qiu, Wei, and
  Zhou}]{hu2017reinforced}
Minghao Hu, Yuxing Peng, Zhen Huang, Xipeng Qiu, Furu Wei, and Ming Zhou. 2018.
\newblock Reinforced mnemonic reader for machine reading comprehension.
\newblock In \emph{Proceedings of IJCAI}.

\bibitem[{Huang and Carley(2018)}]{huang2018parameterized}
Binxuan Huang and Kathleen Carley. 2018.
\newblock Parameterized convolutional neural networks for aspect level
  sentiment classification.
\newblock In \emph{Proceedings of EMNLP}.

\bibitem[{Jakob and Gurevych(2010)}]{jakob2010extracting}
Niklas Jakob and Iryna Gurevych. 2010.
\newblock Extracting opinion targets in a single-and cross-domain setting with
  conditional random fields.
\newblock In \emph{Proceedings of EMNLP}.

\bibitem[{Jiang et~al.(2011)Jiang, Yu, Zhou, Liu, and Zhao}]{jiang2011target}
Long Jiang, Mo~Yu, Ming Zhou, Xiaohua Liu, and Tiejun Zhao. 2011.
\newblock Target-dependent twitter sentiment classification.
\newblock In \emph{Proceedings of ACL}.

\bibitem[{Kim(2014)}]{kim2014convolutional}
Yoon Kim. 2014.
\newblock Convolutional neural networks for sentence classification.

\bibitem[{Lafferty et~al.(2001)Lafferty, McCallum, and
  Pereira}]{lafferty2001conditional}
John Lafferty, Andrew McCallum, and Fernando~CN Pereira. 2001.
\newblock Conditional random fields: Probabilistic models for segmenting and
  labeling sequence data.
\newblock In \emph{Proceedings of ICML}.

\bibitem[{Lee et~al.(2017)Lee, He, Lewis, and Zettlemoyer}]{lee2017end}
Kenton Lee, Luheng He, Mike Lewis, and Luke Zettlemoyer. 2017.
\newblock End-to-end neural coreference resolution.
\newblock \emph{arXiv preprint arXiv:1707.07045}.

\bibitem[{Lee et~al.(2016)Lee, Salant, Kwiatkowski, Parikh, Das, and
  Berant}]{lee2016learning}
Kenton Lee, Shimi Salant, Tom Kwiatkowski, Ankur Parikh, Dipanjan Das, and
  Jonathan Berant. 2016.
\newblock Learning recurrent span representations for extractive question
  answering.
\newblock \emph{arXiv preprint arXiv:1611.01436}.

\bibitem[{Li et~al.(2018)Li, Bing, Lam, and Shi}]{li2018transformation}
Xin Li, Lidong Bing, Wai Lam, and Bei Shi. 2018.
\newblock Transformation networks for target-oriented sentiment classification.
\newblock In \emph{Proceedings of ACL}.

\bibitem[{Li et~al.(2019)Li, Bing, Li, and Lam}]{li2018unified}
Xin Li, Lidong Bing, Piji Li, and Wai Lam. 2019.
\newblock A unified model for opinion target extraction and target sentiment
  prediction.
\newblock In \emph{Proceedings of AAAI}.

\bibitem[{Li and Lam(2017)}]{li2017deep}
Xin Li and Wai Lam. 2017.
\newblock Deep multi-task learning for aspect term extraction with memory
  interaction.
\newblock In \emph{Proceedings of EMNLP}.

\bibitem[{Lin and He(2009)}]{lin2009joint}
Chenghua Lin and Yulan He. 2009.
\newblock Joint sentiment/topic model for sentiment analysis.
\newblock In \emph{Proceedings of CIKM}.

\bibitem[{Liu(2012)}]{liu2012sentiment}
Bing Liu. 2012.
\newblock Sentiment analysis and opinion mining.
\newblock \emph{Synthesis lectures on human language technologies},
  5(1):1--167.

\bibitem[{Liu et~al.(2015)Liu, Joty, and Meng}]{liu2015fine}
Pengfei Liu, Shafiq Joty, and Helen Meng. 2015.
\newblock Fine-grained opinion mining with recurrent neural networks and word
  embeddings.
\newblock In \emph{Proceedings of EMNLP}.

\bibitem[{Mitchell et~al.(2013)Mitchell, Aguilar, Wilson, and
  Van~Durme}]{mitchell2013open}
Margaret Mitchell, Jacqui Aguilar, Theresa Wilson, and Benjamin Van~Durme.
  2013.
\newblock Open domain targeted sentiment.
\newblock In \emph{Proceedings of EMNLP}.

\bibitem[{Pang et~al.(2008)Pang, Lee et~al.}]{pang2008opinion}
Bo~Pang, Lillian Lee, et~al. 2008.
\newblock Opinion mining and sentiment analysis.
\newblock \emph{Foundations and Trends{\textregistered} in Information
  Retrieval}, 2(1--2):1--135.

\bibitem[{Pontiki et~al.(2016)Pontiki, Galanis, Papageorgiou, Androutsopoulos,
  Manandhar, Mohammad, Al-Ayyoub, Zhao, Qin, De~Clercq
  et~al.}]{pontiki2016semeval}
Maria Pontiki, Dimitris Galanis, Haris Papageorgiou, Ion Androutsopoulos,
  Suresh Manandhar, AL-Smadi Mohammad, Mahmoud Al-Ayyoub, Yanyan Zhao, Bing
  Qin, Orph{\'e}e De~Clercq, et~al. 2016.
\newblock Semeval-2016 task 5: Aspect based sentiment analysis.
\newblock In \emph{Proceedings of SemEval-2016}.

\bibitem[{Pontiki et~al.(2015)Pontiki, Galanis, Papageorgiou, Manandhar, and
  Androutsopoulos}]{pontiki2015semeval}
Maria Pontiki, Dimitris Galanis, Haris Papageorgiou, Suresh Manandhar, and Ion
  Androutsopoulos. 2015.
\newblock Semeval-2015 task 12: Aspect based sentiment analysis.
\newblock In \emph{Proceedings of SemEval 2015}.

\bibitem[{Pontiki et~al.(2014)Pontiki, Galanis, Pavlopoulos, Papageorgiou,
  Androutsopoulos, and Manandhar}]{pontiki2014semeval}
Maria Pontiki, Dimitris Galanis, John Pavlopoulos, Harris Papageorgiou, Ion
  Androutsopoulos, and Suresh Manandhar. 2014.
\newblock Semeval-2014 task 4: Aspect based sentiment analysis.
\newblock In \emph{Proceedings of SemEval-2014}.

\bibitem[{Poria et~al.(2016)Poria, Cambria, and Gelbukh}]{poria2016aspect}
Soujanya Poria, Erik Cambria, and Alexander Gelbukh. 2016.
\newblock Aspect extraction for opinion mining with a deep convolutional neural
  network.
\newblock \emph{Knowledge-Based Systems}, 108:42--49.

\bibitem[{Rajpurkar et~al.(2016)Rajpurkar, Zhang, Lopyrev, and
  Liang}]{Rajpurkar16}
Pranav Rajpurkar, Jian Zhang, Konstantin Lopyrev, and Percy Liang. 2016.
\newblock Squad: 100,000+ questions for machine comprehension of text.
\newblock In \emph{Proceedings of EMNLP}.

\bibitem[{Rosenfeld and Thurston(1971)}]{rosenfeld1971edge}
Azriel Rosenfeld and Mark Thurston. 1971.
\newblock Edge and curve detection for visual scene analysis.
\newblock \emph{IEEE Transactions on computers}, (5):562--569.

\bibitem[{Seo et~al.(2017)Seo, Kembhavi, Farhadi, and
  Hajishirzi}]{seo2016bidirectional}
Minjoon Seo, Aniruddha Kembhavi, Ali Farhadi, and Hannaneh Hajishirzi. 2017.
\newblock Bidirectional attention flow for machine comprehension.
\newblock In \emph{Proceedings of ICLR}.

\bibitem[{Shu et~al.(2017)Shu, Xu, and Liu}]{shu2017lifelong}
Lei Shu, Hu~Xu, and Bing Liu. 2017.
\newblock Lifelong learning crf for supervised aspect extraction.
\newblock In \emph{Proceedings of the ACL}.

\bibitem[{Tang et~al.(2016{\natexlab{a}})Tang, Qin, Feng, and
  Liu}]{tang2015effective}
Duyu Tang, Bing Qin, Xiaocheng Feng, and Ting Liu. 2016{\natexlab{a}}.
\newblock Effective lstms for target-dependent sentiment classification.
\newblock In \emph{Proceedings of COLING}.

\bibitem[{Tang et~al.(2016{\natexlab{b}})Tang, Qin, and Liu}]{tang2016aspect}
Duyu Tang, Bing Qin, and Ting Liu. 2016{\natexlab{b}}.
\newblock Aspect level sentiment classification with deep memory network.
\newblock \emph{arXiv preprint arXiv:1605.08900}.

\bibitem[{Vaswani et~al.(2017)Vaswani, Shazeer, Parmar, Uszkoreit, Jones,
  Gomez, Kaiser, and Polosukhin}]{vaswani2017attention}
Ashish Vaswani, Noam Shazeer, Niki Parmar, Jakob Uszkoreit, Llion Jones,
  Aidan~N Gomez, {\L}ukasz Kaiser, and Illia Polosukhin. 2017.
\newblock Attention is all you need.
\newblock In \emph{Proceedings of NIPS}.

\bibitem[{Wang and Jiang(2017)}]{wang2016machine}
Shuohang Wang and Jing Jiang. 2017.
\newblock Machine comprehension using match-lstm and answer pointer.
\newblock In \emph{Proceedings of ICLR}.

\bibitem[{Wang et~al.(2016{\natexlab{a}})Wang, Pan, Dahlmeier, and
  Xiao}]{wang2016recursive}
Wenya Wang, Sinno~Jialin Pan, Daniel Dahlmeier, and Xiaokui Xiao.
  2016{\natexlab{a}}.
\newblock Recursive neural conditional random fields for aspect-based sentiment
  analysis.
\newblock In \emph{Proceedings of EMNLP}.

\bibitem[{Wang et~al.(2016{\natexlab{b}})Wang, Huang, Zhao
  et~al.}]{wang2016attention}
Yequan Wang, Minlie Huang, Li~Zhao, et~al. 2016{\natexlab{b}}.
\newblock Attention-based lstm for aspect-level sentiment classification.
\newblock In \emph{Proceedings of EMNLP}.

\bibitem[{Xu et~al.(2018)Xu, Liu, Shu, and Yu}]{xu2018double}
Hu~Xu, Bing Liu, Lei Shu, and Philip~S Yu. 2018.
\newblock Double embeddings and cnn-based sequence labeling for aspect
  extraction.
\newblock In \emph{Proceedings of ACL}.

\bibitem[{Xue and Li(2018)}]{xue2018aspect}
Wei Xue and Tao Li. 2018.
\newblock Aspect based sentiment analysis with gated convolutional networks.
\newblock \emph{arXiv preprint arXiv:1805.07043}.

\bibitem[{Zhang et~al.(2015)Zhang, Zhang, and Vo}]{zhang2015neural}
Meishan Zhang, Yue Zhang, and Duy~Tin Vo. 2015.
\newblock Neural networks for open domain targeted sentiment.
\newblock In \emph{Proceedings of EMNLP}.

\end{thebibliography}
\bibliographystyle{acl_natbib}

\end{document}